# An online supervised learning algorithm based on triple spikes for spiking neural networks

Guojun Chen[1], Xianghong Lin[2] and Guoen Wang[1*]

*1. School of Urban Design, Wuhan University, Wuhan 430072, PR China*

*2. School of Computer Science and Engineering, Northwest Normal University, Lanzhou 730070, PR China*

**Abstract:** Using precise times of every spike, spiking supervised learning has more effects on complex spatial-temporal pattern than supervised learning only through neuronal firing rates. The purpose of spiking supervised learning after spatial-temporal encoding is to emit desired spike trains with precise times. Existing algorithms of spiking supervised learning have excellent performances, but mechanisms of them still have some problems, such as the limitation of neuronal types and complex computation. Based on an online regulative mechanism of biological synapses, this paper proposes an online supervised learning algorithm of multiple spike trains for spiking neural networks. The proposed algorithm with a spatial-temporal transformation can make a simple direct regulation of synaptic weights as soon as firing time of an output spike is obtained. Besides, it is also not restricted by types of spiking neuron models. Relationship among desired output, actual output and input spike trains is firstly analyzed and synthesized to simply select a unit of pair-spike for a direct regulation. And then a computational method is constructed based on simple triple spikes using this direct regulation. Compared with other learning algorithms, results of experiments show that proposed algorithm has higher learning accuracy and efficiency.

## 1. Introduction

Although algorithms using CSSM have better performances, reasonable explanation about this method is not given. It is only argued that some fuzzy local overlaps are avoided partly. According to an analysis of above learning patterns and methods, problems of GSSM are cognized globally based on a divide of intervals for spike trains in this paper. Then, really effective part of a conversion from offline pattern to online pattern is spatially-temporally transformed to an online simple-direct selection and computation method. However, the unit of pair-spike obtained by this method also has problems. Analyzing firing times of spikes in different intervals, an online supervised learning algorithm with triple spikes based on difference and proportion is constructed in this paper. We denote this method as triple-spike-driven (TSD). The inspiration of triple-spike stems from triple-STDP (TSTDP). TSTSP is a rule of synapse plasticity changed from STDP, where the latest output spike of current output spike at the same spike train is used [34]. Then Lin et al. [35] put it into ReSuMe to obtain an improvement. Being different from existing TSTDP, a new method of triple spikes in this paper is proposed. The rest of this paper is organized as follows. In Section 2, spiking neuron models and network architecture used in this paper are introduced. In Section 3, an online supervised learning algorithm TSD is constructed and analyzed. In section 4, some experiments are carried out to demonstrate learning capabilities of TSD. Then, other spiking supervised learning algorithms are taken into comparisons with TSD respectively. The discussion is presented in section 5. Finally, section 6 makes a conclusion and describes future work.

## 2. Spiking neuron model and network architecture

### 2.1 Spiking neuron model

### 2.2 The architecture of SNNs

## 3. The online supervised learning algorithm: triple-spike-driven

### 3.1 The divide of intervals for spike trains

The spike train $s = \{t^f \in \Gamma : f = 1,2,...,F\}$ represents an ordered sequence of spiking times fired by a spiking neuron in an

[*] Corresponding author.

*E-mail address:* 150251932@qq.com (G. Wang).

interval $T[0,T]$. It is shown formally as:

$$s(t) = \sum_{f=1}^{F} \delta(t - t^f) \quad (4)$$

, where $F$ is the number of spikes, $\delta(t)$ is a Dirac delta function, if $t = 0$, $\delta(x) = 1$ or $\delta(x) = 0$ otherwise. $f$ is $f$th spike firing time. Input spike train can be expressed as $s_i$, actual spike train is $s_a$ and desired spike train is $s_d$.

For one spike at $s_d$ in the supervised layer, membrane potential after firing this spike will change to a setting constant immediately in typical spiking neuronal models. Thus, $s_d$ can be seen as a relative recurrent process, where every interval between two nearest desired output spikes is a step. The divide of intervals for desired spike train is shown in Fig. 2, an interval of one step is called desired time interval (DTI). The first DTI can be expressed as $[0, t_d^1]$, the $n$th spike is $[t_d^{n-1}, t_d^n]$. With the same method, an interval of one step at $s_a$ in output layer is called actual time interval (ATI). The first ATI is expressed as $[0, t_a^1]$, the $n$th spike is $[t_a^{n-1}, t_a^n]$. All time intervals of spike trains are divided based on DTI in this section due to desired spike trains as a symbol of successful learning is related to all spikes in a DTI. Besides, DTI as a basic interval is analyzed to construct supervised learning algorithm based on differences between desired and actual output spikes in one DTI expediently.

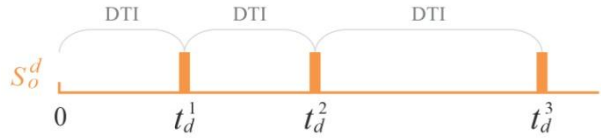

**Fig.2.** The divide of intervals for a desired output spike train. The interval between two nearest spikes at one desired output train is a desired time interval (DTI), while the interval before first spike is start at 0ms.

Spikes of $s_a$ in a DTI can be discussed in two types. In first type, an actual output spike at $s_a$ is produced by input spikes in a DTI at all. For example, firing time of the first actual output spike $t_a^1$ at $s_a$ is earlier than firing time of first desired output spike $t_d^1$ at $s_d$. In this example, an interval of time between $t_d^1$ and $t_a^1$ is called general actual time interval (GATI). When the number of actual output spikes is more than one in a DTI, all these actual output spikes are produced by input spikes in this DTI. Intervals of time between two nearest actual output spikes are also called GATI. In second type, an actual output spike at $s_a$ is not all produced by input spikes in a DTI in the second type. It means that this actual output spike at $s_a$ is fired in next DTI and produced by input spikes in current DTI partly or at all. For example, firing time of $t_a^1$ at $s_a$ is later than $t_d^1$ at $s_d$. In this example, an interval of time between 0 and $t_d^1$ is called special actual time interval (SATI). When an actual output spike of this type is not the first one at $s_a$, the interval between prior actual output spike and current actual output spike also is SATI. If the number of actual output spikes in one DTI is more than one, actual output spikes except first one are all produced by input spikes in this DTI. Intervals of time between two nearest actual output spikes are GATI. Moreover, it should be noted that the first actual output spike in current DTI may not be produced by input spikes of prior DTI, which is similar to the first type. Fig. 3 is an example of divide of intervals for actual spike train $s_a$ with an enough big rate of input spike trains. It can be seen that four ATI are existed at actual output train, with two GATI, $[0, t_a^1]$ and $[t_a^3, t_a^4]$, and two SATI, $[t_a^1, t_a^2]$ and $[t_a^2, t_a^3]$.

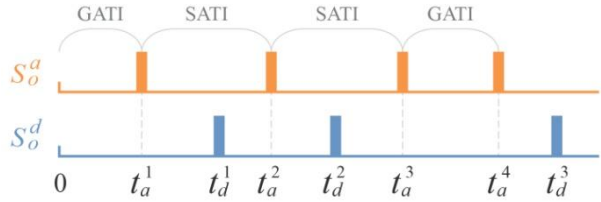

**Fig.3.** The divide of intervals for actual output spike train $s_a$. The interval between two nearest spikes at one actual output spike train is an actual time interval (ATI). The interval before $t_a^1$ is start at 0ms. When two nearest actual output spikes for one ATI are in the same DTI, this ATI is called general actual time interval (GATI), e.g. $[0, t_a^1]$ and $[t_a^3, t_a^4]$. ATI is a special actual time interval (SATI) when two nearest actual output spikes for one ATI are in different DTI, e.g. $[t_a^1, t_a^2]$ and $[t_a^2, t_a^3]$.

Intervals of input spikes at $s_i$ based on $s_d$ and $s_a$ can be divided into three conditions in one DTI. Being different from above intervals of time, intervals of input spikes are used to classify types of input spikes. Condition 1: there is more than one actual output spike in one DTI. In this condition, DTI for input spikes makes up with three parts, first input time interval (FITI) from first desired output spike to first actual output spike, median input time interval (MITI) between first actual output spike and last actual spike and last input time interval (LITI) from last actual output spike to next desired output spike. Input spikes in two parts produce first actual output spike, where part 1 is FITI in current DTI, the other is LITI of prior DTI, and the sum of them equals SATI. Input spikes in MITI produce other actual output spikes except first one. Combined with input spikes in FITI of next DTI, input spikes in LITI of current DTI produce first actual output spike in next DTI. Condition 2: only one actual output spike is fired in one DTI. MITI does not exist in this condition while one DTI is divided into FITI and LITI. Condition 3: there is no one actual spike in one DTI. It can be seen this DTI as one part of FITI in next DTI. All input spikes in this DTI produce first actual output spike of next DTI. An example is shown in Fig.4, where $[t_d^1, t_a^1]$ is a FITI, $[t_a^1, t_a^2]$ and $[t_a^2, t_a^3]$ are MITI, $[0, t_d^1]$ and $[t_a^3, t_d^2]$ are LITI.

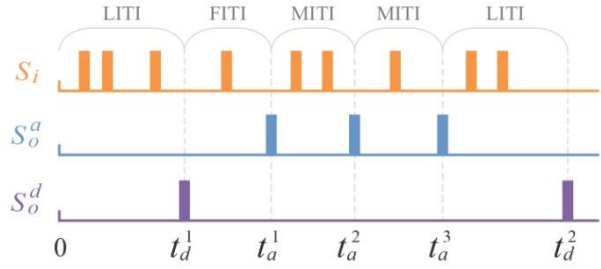

**Fig.4.** The divide of intervals for an input spike train $s_i$. Based on a desired output spike train and an actual output spike train, the intervals of input spike train are divided in DTI. In a DTI, the interval from first desired output spike to first actual output spike is first input time interval (FITI), e.g. $[t_d^1, t_a^1]$; the interval between first actual output spike and last actual output spike is median input time interval (MITI), e.g. $[t_o^1, t_a^2]$ and $[t_a^2, t_a^3]$; and the interval from last actual output spike to current desired output spike is last input time interval (LITI), e.g. $[t_a^3, t_d^2]$ and $[0, t_d^1]$ which equals to a DTI.

### 3.2 A simple direct selection and computation method of spikes

Selection and computation method of spikes in many traditional algorithms such as ReSuMe, SPAN, PSD, STKLR and etc. are all the same[31]. In order to calculate distances between actual and desired output spikes to regulate weights, connections between input and output spikes which all have relations with synaptic weights are used. Due to input spikes in one DTI (or ATI) produce a desired output spike (or an actual output spike), all input spikes in DTI (or ATI) are calculated generally when one output spike comes. There are two time intervals used, one is between an input spike and a desired output spike, the other is between an input spike and an actual output spike. Differences between the two time intervals with conversion and transformation or not are calculated to regulate a weight. As a result, every input spike before both last desired output spike and last actual output spike is calculated in twice, one time of twice calculations is to increase the value of synaptic weight and the other one is to decrease this value. Input spike used once is in last DTI (or ATI). The basic rule of this method is that synaptic weight will be decreased if an actual output spike is early than a desired output spike while this weight is increased if an actual output spike is early than a desired output spike [23].

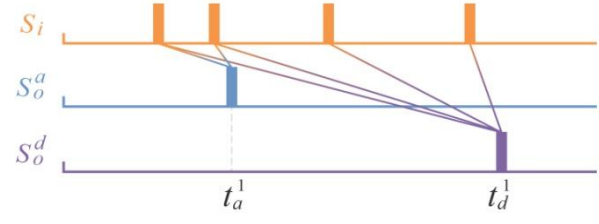

**Fig.5.** The traditional selection and computation method of spikes. All input spikes in a DTI (or ATI) are calculated for the desired output spike (or actual output spike), resulting from that these input spikes in one DTI (or ATI) produce the desired output spike (or actual output spike).

The accuracy and efficiency of above rule are proved by experiments of many algorithms. However, problems of this rule still exist in actual calculations. As it can be seen in Fig.5, input spikes before actual output spike $t_a^1$ in GATI are used to decrease synaptic weight while input spikes before a desired output spike $t_d^1$ in this DTI is calculated to increase this weight. Due to $t_a^1$ is before than $t_d^1$, this weight should be decreased. However, this weight can be able to increase by traditional method, which is opposite to correct regulation that supervised layer desired. Because input spikes before $t_a^1$ in FITI are used twice, one is for increasing and the other one is for decreasing, while input spikes after $t_a^1$ but before $t_d^1$ in LITI are used just once for decreasing in online learning. Online pattern of supervised learning using this rule can be thought overlaps locally [31, 33], but a globally basic reason is that numbers of input spikes used in two parts have difference. In this example of Fig.5, the number of input spikes used for $t_d^1$ is more than the number for $t_a^1$ in DTI. In other words, using common number of input spikes to regulate synaptic weight is necessary for supervised learning, which can make a common condition for two parts of supervised learning. With the same reason, there is also a mistake when $t_d^1$ is before than $t_a^1$.

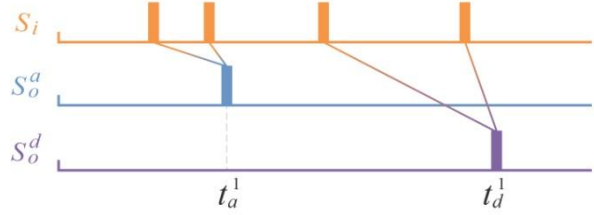

**Fig.6.** The simple direct selection and computation method of spikes. Every input spike in each DTI or ATI is only used once to calculate regulated value of a synaptic weight directly, which can be seen as a result of two regulations using a unit of spikes, an increment made by a desired output spike and a decrement made by an actual output spike.

Actually, the rule of many algorithms used in online learning is same with in offline learning. In offline learning, common number of input spikes is used in two parts of supervised learning. Here, a common WH rule that is also known as Delta rule is used to regulate synaptic weights [25, 27, 29, 31]. The changed value of $i$th synaptic weight $\Delta w_i$ from presynaptic neuron to postsynaptic neuron is regulated as follows:

$$\Delta w_i = \eta s_i \left[ s_d - s_a \right] \tag{5}$$

, where $s_i$ is an input spike train transferred through synapse $i$, $s_a$ is an actual output spike train, $s_d$ is a desired output spike train. At $t$ time, $\Delta w_i(t)$ is expressed as:

$$\Delta w_i(t) = \eta s_i(t) \left[ s_d(t) - s_a(t) \right] \tag{6}$$

, where $s_i(t)$ is an input train, $s_a(t)$ is an actual output train, $s_d(t)$ is a desired output train. Formalized Eq. (4) can be represented as:

$$p_i(t) = \sum_f^F \kappa(t - t^f) \tag{7}$$

, where $\kappa(t)$ is a kernel function such as Gaussian, Laplace, alpha and exponential kernels. Thus, Eq. (6) is expressed as:

$$\Delta w_i(t) = \eta \kappa(t - t_i^f) \left[ \kappa(t - t_d^m) - \kappa(t - t_a^n) \right] \tag{8}$$

, where $t_i^f$ denotes the firing time of $f$th input spike, $t_d^m$ denotes the firing time of $m$th desired output spike, $t_a^n$ is the firing time of $n$th actual output spike. The $m$th desired output is denoted as the first desired output spike after $f$th input spike, the $n$th desired output is denoted as the first desired output spike after $f$th input spike. From Eq. (8), it can be known at $t$ time that

$$\Delta w_i(t) \begin{cases} > 0, & \text{if } \kappa(t - t_d^m) - \kappa(t - t_a^n) > 0 \\ = 0, & \text{if } \kappa(t - t_d^m) - \kappa(t - t_a^n) = 0 \\ < 0, & \text{if } \kappa(t - t_d^m) - \kappa(t - t_a^n) < 0 \end{cases} \tag{9}$$

. Based on Eq. (9), synaptic weight at $t$ time will be increased when $m$th desired output spike is before than $n$th actual output spike. This weight will be decreased conversely. Thus, the function of every input spike is certain for synaptic weight when firing times of actual output spikes are obtained in a DTI. Besides, input spikes for a corresponding desired output spike and an actual output spike can be ensured only after firing times of actual output spikes are obtained in one DTI. Firing time $t_a^n$ is in a DTI with $t_d^m$ when an input spike $t_i^f$ is in FITI or MITI. In this condition, the distance between $t_a^n$ and $t_i^f$ is shorter than the distance between $t_d^m$ and $t_i^f$. $\Delta w_i(t)$ is decreased. Firing time $t_a^n$ is also in a DTI with $t_d^m$ when an input spike $t_i^f$ is in LITI. In this condition, the distance between $t_a^{n+1}$ and $t_i^f$ is longer than the distance between $t_d^m$ and $t_i^f$ when $t_a^{n+1}$ exists, $\Delta w_i(t)$ is increased. The condition that firing time $t_a^n$ is not in the DTI with $t_d^m$ is same with the condition that an input spike $t_i^f$ is in LITI. $\Delta w_i(t)$ in this condition is increased. Based on above three conditions, it can be concluded that $\Delta w_i(t)$ is increased when the first output spike is actual output spike while $\Delta w_i(t)$ is increased when the first output spike is actual output spike. Firing time of an input spike can make a decision to increase or decrease synaptic weight when the first output spike after this input spike is known.

When the selection method is known, the rest question is how to make a computation. No matter how long the distance between $t_d^m$ and $t_a^n$ is in a DTI, it is 0 when supervised learning is successful. It means that the distance between $t_a^n$ and $t_i^f$ equals to the distance between $t_d^m$ and $t_i^f$. Due to $t_d^m$ and $t_i^f$ are known before supervised learning, $\kappa(t - t_d^m)$ is a constant. $\kappa(t - t_a^n)$ approaches $\kappa(t - t_d^m)$ when the distance between $t_d^m$ and $t_a^n$ from approaches 0. This is usually an iterative process. In this process, input spikes in FITI and MITI are used for $\kappa(t - t_d^m)$ firstly when the learning is unsuccessful. Then, they are used for $\kappa(t - t_a^n)$ to judge the success of learning when $t_a^n$ equals $t_d^m$. For example, input spikes in GATI before $t_a^1$ are used to increase synaptic weight in Fig.5. Then, these spikes are used to decrease synaptic weight when $t_a^1$ closes to $t_d^1$. Distances they used are equivalent. From the view of a whole process, input spikes in FITI and MITI should not be used for $\kappa(t - t_d^m)$ at each epoch. At the same time, input spikes in DTI without any actual output spike and LITI should not be used for $\kappa(t - t_d^m)$ at each epoch. The effect of them is cancellable temporally. In this way, every input spike needed is used only once to calculate a difference directly at each epoch. In order to construct a direct computation based on an output spike, input spikes only before this output spike are taken into account. This direct computation can also be regard as a result of every input spike calculated twice. It also ensures that the number of input spikes used for desired output spikes and the number of input spikes used for actual output spikes are same. Hence, Eq. (8) as a formula of online learning is made a spatial-temporal change to

$$\Delta w_i(t) = \eta \cdot \alpha(t) \sum_{t^* \in (t^p, t]}^{t} \kappa(t - t^*) \tag{10}$$

$$\alpha(t) = \begin{cases} 1, & if\ t - t_d^m = 0, t - t_a^n \neq 0 \\ -1, & if\ t - t_d^m = 0, t - t_a^n \neq 0 \\ 0, & otherwise \end{cases} \tag{11}$$

$$t^p = \begin{cases} t_d^{m-1}, & if\ t_a^{n-1} < t_d^{m-1}, m > 1, n > 1 \\ t_a^{n-1}, & if\ t_a^{n-1} \geq t_d^{m-1}, m > 1, n > 1 \\ t_d^{m-1}, & if\ m > 1, n = 1 \\ t_a^{n-1}, & if\ m = 1, n > 1 \\ 0, & if\ m = 1, n = 1 \end{cases} \tag{12}$$

. According to above analyses, a simple direct selection and computation method of spikes in online pattern can be designed. The method can be described that one input spike is only used once to regulate synaptic weight depending on first output spike after it at one learning epoch. The input spike and its output spike constitute a unit of pair-spike. Input spikes after all output spikes in neither DTI nor ATI are not taken into calculation. Because no one output spike after them exists or they have no effect on output spikes. It is shown in Fig.6 that input spikes in FITI before $t_a^1$ should be used to increase synaptic weight in the example of Fig.5, while input spikes after $t_a^1$ but before $t_d^1$ in LITI should be used to decrease synaptic weight. Besides, one condition that the number of spikes at an actual output train does not equal to number of spikes at a desired output train can occur. It usually makes desired and actual output spikes could not be one to one correspondence [31]. Being different from a method in [31], this condition here can be seen that firing time of the output spike corresponding at the other output spike train is positive infinity.

### 3.3 Triple-spikes-driven based on proportion

Regulation rule in Eq. (10) is not holonomic. In Fig.6, when the length between an input spike and actual output spike of FITI is a constant in one DTI, the decrement of weight could not be varied with changed length of FITI; when the length between an input spike and actual output spike of LITI is a constant in one DTI, the increment of weight could not be varied with changed length of LITI. In other words, regulation rule in Eq. (10) lacks a function of two nearest output spikes for only intervals of input spikes and output spikes are calculated. The method in section 3.2 is only a part of real direct online learning. For this problem, a proportion of two distances based on TSTDP is used. TSTDP is a variant of STDP where latest spike of current output spike in one output spike train is taken into account. STDP is an effective learning rule for spiking neural networks. Some excellent research results of STDP learning algorithms can still be obtained along with developments of SNNs [15]. Besides, algorithms based on STDP also only depend on firing times, which is not restricted to specific dynamic of spiking neuron models. Although STDP is only a rule for unsupervised learning in the beginning, extensions embody relationships of input, actual output and desired output spikes for supervision [21]. For example, ReSuMe can make neurons learn to emit multiple spikes, which has an inevitable biological basis and practical application [23, 24]. The formation of STDP is

$$\Delta w = \begin{cases} \Delta w^+ = A^+ \exp\left(\dfrac{-\Delta t}{\tau^+}\right) & if\ \Delta t > 0 \\ \Delta w^- = A^- \exp\left(\dfrac{-\Delta t}{\tau^-}\right) & if\ \Delta t \leq 0 \end{cases} \quad (13)$$

, where $\Delta w$ denotes changed value of synaptic weight. $\Delta t = t_{post} - t_{pre}$ denotes the difference of timing between a presynapstic spike and a postsynaptic spike. $t_{pre}$ is firing time at presynaptic and $t_{post}$ is firing time at postsynaptic. $A^+$ and $A^-$ denote parameters of potentiation and depression respectively. Besides, $\tau^+$ and $\tau^-$ are constants. The formation of TSTDP is

$$\Delta w = \begin{cases} \Delta w^+ = A^+ \exp\left(\dfrac{-\Delta t}{\tau^+}\right)\left[A_2^+ + A_3^+ \exp\left(\dfrac{-\Delta t_2}{\tau_y}\right)\right] & if\ \Delta t > 0 \\ \Delta w^- = A^- \exp\left(\dfrac{-\Delta t}{\tau^-}\right)\left[A_2^- + A_3^- \exp\left(\dfrac{-\Delta t_3}{\tau_y}\right)\right] & if\ \Delta t \leq 0 \end{cases} \quad (14)$$

, where $\Delta w$ equals to $\Delta w^+$ when $t = t_{post}$ and $\Delta w$ equals to $\Delta w^-$ when time $t = t_{pre}$. $\Delta t_1 = t_{post(n)} - t_{pre(n)}$, $\Delta t_2 = t_{post(n)} - t_{post(n-1)}$ and $\Delta t_3 = t_{pre(n)} - t_{pre(n-1)}$ denote differences of timing between presynapstic spikes and postsynaptic spikes. $A_2^+$, $A_2^-$, $A_3^+$ and $A_3^-$ denote parameters of potentiation and depression, $\tau_y$ is a constant. Using the distance between two nearest output spikes in one output spike train, distribution of values in TSTDP can be more approximate to biological data [34]. Moreover, TSTDP can also be reflex to BCM-like rules, which has better performances in memoritive circuits [36, 37]. Being different from Eq. (14), a proportion is used to replace a difference at the right part. This proportion of distance also has ability to learning, which is same with the difference of distances. Latest spike of original output train in TSTDP is altered to be the nearest output spike before an input spike. Therefore, these distances are one input spike with their nearest output spikes. An example of TSTDP based on the proportion is shown in Fig.7, which shows that two nearest output spikes of the third input spike are $t_a^1$ and $t_d^1$. Because only input spikes before one output spike for this output spike are used in online pattern, the second part of Eq. (14) rule is unused. Eq. (13) can be seen as a simple type of kernel $\kappa()$, so the formation of an online supervised learning algorithm based on triple spikes can be expressed formally as:

$$\Delta w_i(t) = \eta \cdot \alpha(t) \sum_{t^* \in (t^p, t]} \left\{ \kappa\left(t^* - t_i^f\right)\left[A_2^+ + A_3^+ \exp\left(\dfrac{-(t-t^*)}{\tau_y(t^*-t^p)}\right)\right]\right\} \quad (15)$$

. To avoid the problem of Eq. (13) directly, alone part of traditional regulation is removed by setting parameter $A_2^+$ to 0. The formation of Eq. (15) is changed to:

$$\begin{aligned}\Delta w_i(t) &= \eta \cdot \alpha(t) \sum_{t^* \in (t^p, t]} \left\{ \kappa\left(t^* - t_i^f\right)\left[A_3^+ \exp\left(\dfrac{-(t-t^*)}{\tau_y(t^*-t^p)}\right)\right]\right\} \\ &= \eta \cdot A_3^+ \cdot \alpha(t) \sum_{t^* \in (t^p, t]} \left\{ \kappa\left(t^* - t_i^f\right)\left[\exp\left(\dfrac{-(t-t^*)}{\tau_y(t^*-t^p)}\right)\right]\right\}\end{aligned} \quad (16)$$

. As a result, $A_3^+$ can be seen as a constant combined with $\eta$ by setting $A_3^+ = 1$ in each regulation. Then, the formation of Eq. (16) is changed to:

$$\Delta w_i(t) = \eta \cdot \alpha(t) \sum_{t^* \in (t^p, t]} \left\{ \kappa(t^* - t_i^f) \exp\left(\frac{-(t-t^*)}{\tau_y(t^* - t^p)}\right) \right\} \qquad (17)$$

. Eq. (17) uses a difference and a proportion, which can also be seen as an algorithm with a product of two kernels. With a unit of triple spikes based on a difference and a proportion, this algorithm can be called triple-spike-driven (TSD).

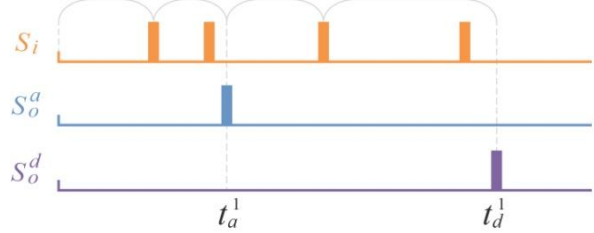

**Fig.7.** The simple direct selection method of triple spikes

## 4. Results of simulations

Several experiments in this section are carried out to demonstrate learning capabilities of TSD. Firstly, TSD is applied to demonstrate capabilities in supervised learning of spatial-temporal spike trains. Then factors that may exert influences on learning performance are analyzed, such as learning epoch, *LR*, number of synaptic input neurons, length of simulative time of spike trains and firing rate of spike trains. Furthermore, in order to know whether TSD has better learning performances than other algorithms or not, comparisons of TSD and some other algorithms are carried out.

Table 1
Value of SRM neurons' parameters

| Parameters | $\tau$ | $\tau_R$ | $t_R$ | $\upsilon$ |
|---|---|---|---|---|
| Value | 7ms | 80ms | 1ms | 1mV |

A spiking neural network (SNN) with architecture of two neuronal layers is used in simulations, where neurons are above described SRM. Due to experiments are used to simulate SNNs by computer programs with discrete running time, spiking neurons in these programs have to progress in a discrete mode. In this paper, the discrete precision of spiking neuron as a parameter is 0.1 ms. Besides, values of parameters of SRM applied in these experiments are listed in Table1: time constant of postsynaptic potential $\tau$ =7ms, time constant of refractory period $\tau_R$=80ms, neuronal threshold $\upsilon$=1mV, and length of absolute refractory period $t_R$=1ms. Many computational methods can be used for $\kappa()$. Here, a Laplace kernel as an example is used. Parameters of TSD are $\tau^+$=7ms and $\tau_y$=7ms. Then, the mathematical representation of TSD with Laplace kernel is given as:

$$\begin{aligned}\Delta w_i(t) &= \eta \cdot \alpha(t) \sum_{t^* \in (t^p, t]} \left\{ \exp\left(\frac{-(t-t^*)}{\tau^+}\right) \exp\left(\frac{-(t-t^*)}{\tau_y(t^* - t^p)}\right) \right\} \\ &= \eta \cdot \alpha(t) \sum_{t^* \in (t^p, t]} \left\{ \exp\left(\frac{-(t-t^*)}{\tau^+} + \frac{-(t-t^*)}{\tau_y(t^* - t^p)}\right) \right\}\end{aligned} \qquad (18)$$

In order to evaluate learning performances quantitatively and directly, correlation-based measure *C* used in many papers [18, 23, 25, 31, 32, and 35] is set to measure the distance between desired and actual output spike trains. The computational formula of *C* after each learning epoch is

$$C = \frac{v_a \cdot v_d}{|v_a||v_d|} \qquad (19)$$

, where $v_a$ and $v_d$ are vectors representing of actual and desired output spike trains with a Gaussian low-pass filter, $v_a \cdot v_d$ is an inner product of $v_a$ and $v_d$, and $|v_a|$, $|v_d|$ are vector modules respectively. Successful learning is accomplished when *C* equals to 1. Besides, an efficiency function of epochs is also constructed. Maximum *C* only could be replaced when *C* is greater than original one and average increment of every epoch from original one to new one is greater than $10^{-5}$.

### 4.1 Two learning examples of TSD

In this section, two learning examples are carried on to prove capabilities of TSD in online pattern. One is a simple example and the other is a complex example. The aim of these learning is through training SNNs to emit desired spatial-temporal spike trains. Learning process, learning accuracy and synaptic weights are investigated in every example.

### 4.1.1 A simple example of TSD
### 4.1.2 A complex example of TSD
### 4.2. Influences of learning factors

With more complex SNN, learning results of the complex example are worse than that of simple one in section 4.1. Complexities of the second example are shown in different values of factors. So these factors may have influences on learning capabilities of TSD respectively. In this section, influences of learning factors such as iterative learning epoch, number of input neurons, *LR*, length of spike trains and Poisson rate of spike trains are investigated on learning results.

### 4.2.1 Influence of iterative learning epoch
### 4.2.2 Influence of learning rate
### 4.2.3 Influence of number of input neuron
### 4.2.4 Influence of length of spike trains
### 4.3 Comparison of TSD and other algorithms

TSD has an ability to learn effectively, but other existing algorithms can also have abilities to learn well. To demonstrate the advantages of TSD, some algorithms including the traditional offline pattern of TSD are used to compare with TSD in this section. They are such as ReSuMe, SPAN and nSTK. Offline pattern is comprehensively compared for different patterns. Traditional offline pattern of TSD is an offline learning method without triple-spike, which can be transformed from Eq. (5). It is also one form of STKLR [29]. A reason why other algorithms are selected to compare is that they are different types of typical algorithms. ReSuMe is based on synapse plasticity; SPAN uses convolutions of spike trains; nSTK is an algorithm of kernel functions for spike trains; Supervised learning algorithms based on gradient descent are not used for they have been developed mostly for simple neuron models lonely, which require analytical expression of state variables and cannot be applied to various neuron models [32]. Besides, some algorithms such as FILT are not used to compare for they are only very adaptive to one type of neuronal models. As a result, algorithms compared in this paper are not limited by neuronal models and adaptive to all spiking models. Compared experiments of TSD and other algorithms in online pattern can be taken with transitional length and Poisson rates of spike trains respectively. Because length and Poisson rates of spike trains have obvious effects on accuracy and efficiency of TSD in section 4.2.

### 4.3.1 Comparison of TSD and traditional offline pattern
### 4.3.2 Comparison of TSD and other algorithms with transitional length of spike trains

Table 2
Learning results of algorithms with transitional length of spike trains based on common Poisson rates

| Algorithm | 200ms | | | 400ms | | | 600ms | | | 800ms | | | 1000ms | | | 1200ms | | |
|---|---|---|---|---|---|---|---|---|---|---|---|---|---|---|---|---|---|---|
| | LR | C | Epoch | LR | C | Epoch | LR | C | Epoch | LR | C | Epoch | LR | C | Epoch | LR | C | Epoch |
| ReSuMe | 0.00220 | 0.948 | 1210 | 0.00140 | 0.847 | 1918 | 0.00100 | 0.768 | 1717 | 0.00080 | 0.716 | 1893 | 0.00060 | 0.646 | 2320 | 0.00050 | 0.607 | **2315** |
| SPAN | 0.00024 | 0.887 | 1434 | 0.00008 | 0.801 | 1920 | 0.00006 | 0.692 | 2554 | 0.00004 | 0.614 | 2978 | 0.00003 | 0.587 | 3032 |
| nSTK | 0.00245 | 0.941 | 1036 | 0.00130 | 0.849 | **1376** | 0.00095 | 0.769 | **1663** | 0.00075 | 0.704 | **1889** | 0.00060 | 0.647 | **2129** | 0.00055 | 0.608 | 2407 |
| TSD | 0.00190 | **0.978** | **1022** | 0.00100 | **0.935** | 2067 | 0.00075 | **0.887** | 2239 | 0.00060 | **0.837** | 2417 | 0.00055 | **0.781** | 2839 | 0.00050 | **0.749** | 2830 |

Table 3
Learning results of algorithms with transitional length of spike trains based on different Poisson rates

| Algorithm | 200ms | | | 400ms | | | 600ms | | | 800ms | | | 1000ms | | | 1200ms | | |
|---|---|---|---|---|---|---|---|---|---|---|---|---|---|---|---|---|---|---|
| | LR | C | Epoch | LR | C | Epoch | LR | C | Epoch | LR | C | Epoch | LR | C | Epoch | LR | C | Epoch |
| ReSuMe | 0.051 | 0.947 | 747 | 0.027 | 0.863 | **477** | 0.022 | 0.774 | 1063 | 0.0142 | 0.719 | 822 | 0.0125 | 0.638 | **465** | 0.0105 | 0.607 | **785** |
| SPAN | 0.0036 | 0.895 | **543** | 0.0016 | 0.796 | 949 | 0.001 | 0.720 | 1063 | 0.0008 | 0.668 | 1289 | 0.00065 | 0.619 | 1261 | 0.00045 | 0.574 | 1289 |
| nSTK | 0.053 | 0.95 | 611 | 0.03 | 0.870 | 561 | 0.024 | 0.775 | **492** | 0.0151 | 0.697 | **750** | 0.014 | 0.657 | 553 | 0.0105 | 0.602 | 1101 |
| TSD | 0.0385 | **0.981** | 713 | 0.025 | **0.953** | 892 | 0.018 | **0.893** | 1072 | 0.0138 | **0.836** | 1015 | 0.0122 | **0.789** | 890 | 0.0100 | **0.761** | 1026 |

### 4.3.3 Comparison of TSD and other algorithms with transitional Poisson rate of spike trains

Table 4
Learning results of algorithms with transitional Poisson rates of spike trains based on common Poisson rates

| Algorithm | 50Hz | | | 100 Hz | | | 150 Hz | | | 200 Hz | | | 250 Hz | | | 300 Hz | | |
|---|---|---|---|---|---|---|---|---|---|---|---|---|---|---|---|---|---|---|
| | LR | C | Epoch | LR | C | Epoch | LR | C | Epoch | LR | C | Epoch | LR | C | Epoch | LR | C | Epoch |
| ReSuMe | 0.00220 | 0.955 | 1079 | 0.00140 | 0.847 | 1918 | 0.00100 | 0.783 | **1201** | 0.00080 | 0.744 | **1299** | 0.00060 | 0.713 | **1515** | 0.00050 | 0.712 | **1475** |
| SPAN | 0.00024 | 0.901 | 1301 | 0.00008 | 0.801 | 1920 | 0.00006 | 0.709 | 2130 | 0.00005 | 0.659 | 1864 | 0.00004 | 0.653 | 2056 | 0.00003 | 0.641 | 2551 |
| nSTK | 0.00245 | 0.951 | 1218 | 0.00130 | 0.849 | **1376** | 0.00095 | 0.788 | 1585 | 0.00075 | 0.753 | 1825 | 0.00060 | 0.723 | 1943 | 0.00055 | 0.710 | 1586 |
| TSD | 0.00190 | **0.983** | 1234 | 0.00100 | **0.935** | 2067 | 0.00075 | **0.884** | 1700 | 0.00060 | **0.838** | 1347 | 0.00055 | **0.808** | 1553 | 0.00050 | **0.797** | 1494 |

Table 5
Learning results of algorithms with transitional Poisson rates of spike trains based on different Poisson rates

| Algorithm | 20Hz | | | 40Hz | | | 60Hz | | | 80Hz | | | 100Hz | | | 120Hz | | |
|---|---|---|---|---|---|---|---|---|---|---|---|---|---|---|---|---|---|---|
| | LR | C | Epoch | LR | C | Epoch | LR | C | Epoch | LR | C | Epoch | LR | C | Epoch | LR | C | Epoch |
| ReSuMe | 0.027 | 0.863 | **477** | 0.0083 | 0.846 | **1218** | 0.0035 | 0.848 | 1461 | 0.0018 | 0.855 | 1709 | 0.00140 | 0.847 | 1918 | 0.0012 | 0.841 | 1891 |
| SPAN | 0.0016 | 0.796 | 949 | 0.0006 | 0.785 | 1530 | 0.0003 | 0.804 | 1652 | 0.00013 | 0.789 | 1828 | 0.00008 | 0.801 | 1920 | 0.00005 | 0.792 | 2056 |
| nSTK | 0.03 | 0.870 | 561 | 0.0085 | 0.844 | 1287 | 0.0038 | 0.857 | **1403** | 0.0020 | 0.852 | **1689** | 0.00130 | 0.849 | **1376** | 0.0001 | 0.843 | **1572** |
| TSD | 0.025 | **0.953** | 892 | 0.0080 | **0.936** | 1362 | 0.0032 | **0.939** | 1526 | 0.0015 | **0.941** | 1717 | 0.00100 | **0.935** | 2067 | 0.00075 | **0.929** | 2010 |

## 5. Discussion
## 6. Conclusion and future work

Firing time of spikes can reflected learning results of spiking supervised learning. Selection of spikes used to direct computation is related to evolutional direction. Common number of spikes to compare at each part of the computation is necessary to online learning based on DTI. An online supervised learning algorithm TSD based on triple spikes with spiking times for SNNs is presented in this paper. Combining advantages of simple direct computation of triple spikes and online pattern, this algorithm regulates synaptic weight directly after a spatial-temporal transformation at firing times of output spikes during running process. TSD is shown as an effective and efficient supervised learning algorithm, which make SNNs emit actual output spikes at desired times almost.

**Acknowledgments**

Firstly, we thank editors and reviewers for this manuscript. Then we thank members of Urbaneuron lab for discussions of this manuscript; we are also grateful to Ziyi Chen for designing images. Funding: The work is supported by the National Natural Science Foundation of China under Grants nos. 51878516. Competing interests: The authors declare no conflict of interest.